\documentclass[journal]{IEEEtran}
\usepackage{amsmath,graphicx,amsfonts,epsfig,epstopdf,caption,subcaption,mathrsfs}

\begin{document}

\title{Knowledge Transfer Pre-training}

\author{Zhiyuan Tang,
       {Dong Wang},
       {Yiqiao Pan},
       {Zhiyong Zhang}

\thanks{Zhiyuan Tang, Dong Wang, Yiqiao Pan, Zhiyong Zhang are with the center for speech and language technology (CSLT),  Research Institute of Information Technology at Tsinghua University and the Division of Technical Innovation and Development, Tsinghua National Laboratory for Information Science and Technology. Zhiyuan Tang is also with Chengdu Institute of Computer Applications at Chinese Academy of Sciences and University of Chinese Academy of Sciences. Yiqiao Pan is also with School of Information \& Communication Engineering, Beijing University of Posts and Telecommunications. E-mails: tangzy@cslt.riit.tsinghua.edu.cn; wangdong99@mails.tsinghua.edu.cn; \{panyq, zhangzy\}@cslt.riit.tsinghua.edu.cn. }}

%



\maketitle

\begin{abstract}

Pre-training is crucial for learning deep neural networks. Most of existing pre-training methods train simple models (e.g., restricted Boltzmann machines) and then stack them layer by layer to form the deep structure. This layer-wise pre-training has found
strong theoretical foundation and broad empirical support. However, it is not easy to employ such method to pre-train models
without a clear multi-layer structure, e.g., recurrent neural networks (RNNs). This paper
presents a new pre-training approach based on knowledge transfer learning. In contrast to the layer-wise
approach which trains model components incrementally, the new approach trains the entire model
as a whole but with an easier objective function. This is achieved by utilizing soft targets produced by a prior
trained model (teacher model). Compared to the conventional layer-wise methods,
this new method does not care about the model structure, so can be used to pre-train very complex models.
Experiments on a speech recognition task demonstrated that with this approach, complex RNNs can be
well trained with a weaker deep neural network (DNN) model. Furthermore, the new method can be
combined with conventional layer-wise pre-training to deliver additional gains.

\end{abstract}


\IEEEpeerreviewmaketitle
\section{Introduction}

Deep learning has gained significant success in a wide range of applications, for example, automatic speech recognition (ASR)~\cite{dnn:dahl11:307,dnn:hinton12:141}. Typical deep
models used in ASR include deep neural networks (DNNs)~\cite{dahl2012context,deng2014} and recurrent neural networks (RNNs)~\cite{graves2014towards,sak2014long,weng2014recurrent,yu2015recurrent}. The success of the deep models is largely attributed
to various pre-training approaches that alleviate the under-fitting and over-fitting problem that had hindered
the development of complex neural models for a long time. Most of the well-known pre-training methods are layer-wise, which train simple
models and then stack them layer by layer to form the deep structure. This pre-training is mostly unsupervised, and is
usually followed by a fine-tuning step which refines the model in a supervised fashion. Two popular pre-training approaches are based on
restricted Boltzmann machines (RBMs)~\cite{hinton2006reducing,dahl2012context} and auto-associators~\cite{bengio2007greedy} respectively.

The basic idea of layer-wise pre-training is to divide the hard deep learning task into easier tasks
of training simpler shallow models. A theoretical analysis for its role in deep model training was presented by Bengio~\cite{bengio2007greedy},
and a through empirical analysis was provided by Erhan et al.~\cite{erhan2010does}.
These studies show that the layer-wise pre-training plays a role of regularization that locates the model
to a `good' place in the parameter space so that the succeeding supervised training (aka fine-tuning) is
easy to find a good local minimum. Recently, the effectiveness of
layer-wise pre-training is proved by Paul et al. using the group theory~\cite{pauldoes}. In ASR, Yu and
colleagues reported that a layer-wise discriminative pre-training can obtain similar performance as
the layer-wise unsupervised pre-training does~\cite{yu2010roles,seide2011feature}.
It is now widely accepted that the layer-wise pre-training makes its contribution in two ways:
(1) it can discover hierarchical patterns by which invariant high-level feature can be obtained;
(2) it can initialize deep models in a health state so that the supervised training can be conducted more effectively.

Although theoretically sound and empirically effective, the layer-wise pre-training is limited to multi-layer models.
For models without a clear layer-wise structure, they can not be easily pre-trained by the existing methods. As an example,
the RNN model does not involve a clear layer-wise structure and the model is complicated by the hidden-hidden connections.
To pre-train this model, either an ad-hoc treatment is required or a special pre-training model needs designing.
For example, Vinyals et al. \cite{vinyals2012revisiting} proposed a two-stage approach: in the first stage,
the hidden-hidden connections are cut off and only the forward paths are trained, and in the second stage, the
entire network is optimized. This approach is obviously suboptimal since the recurrent path is not pre-trained together with the
forward path.
Pasa and colleagues~\cite{pasa2014pre} constructed a linear autoencoder on sequential data to pre-train the RNN model.
This linear autoencoder model exactly matches the RNN structure so that all the parameters can be jointly pre-trained.
Following the same idea, Boulanger-Lewandowski et al.~\cite{boulanger2012modeling} proposed a recurrent temporal RBM
model to match the RNN structure. These task-specific pre-training models need to be specifically designed,
which is certainly not ideal. Moreover, if the target model is complex, e.g., with cross-layer
connections, it would be difficult to design an appropriate pre-training model, and training such a model by
unsupervised learning is often a prohibit task.

This paper presents a simple yet powerful pre-training approach based on knowledge transfer, which is largely motivated by the logit matching approach from Ba et al.~\cite{ba2014deep} and the dark knowledge distiller model from Hinton and colleagues~\cite{hinton2014distilling}. The basic idea is that a well-trained model involves rich knowledge and can be used to guide the training of other models. In Ba and Hinton's work~\cite{ba2014deep,hinton2014distilling}, this idea was applied to learn simple models from complex models or model ensembles~\cite{ba2014deep,hinton2014distilling}.
In ASR, Li et al. has applied the same idea to train small DNNs from a large and complex DNN~\cite{li2014learning}.
We show in this paper that knowledge transfer is a general approach and can be used in a very different way. Instead of learning simple models from
complex models, it can be used to pre-train complex models using a simpler model. Specifically, it is possible to train a simple model and then
use this model as a teacher to guide the training of a complex model (child model) that is normally difficult to accomplish. This teacher model might be rather weak, but it is sufficient to direct the child model where to go. Once the teacher model helps the child model reach a reasonable place in
the parameter space, the child model can learn by itself and finally finds a good local optimum, delivering a performance even better than
the teacher model.

This weak teacher strategy is rather different from the idea of logit matching and dark knowledge distillation proposed in~\cite{ba2014deep,hinton2014distilling}. The teacher model plays a role of `supervisor' instead of a `teacher', and the
teaching process is essentially a pre-training. The self-learning of the child model, correspondingly, is a fine-tuning.
In fact, the teaching processing (pre-training) is the same as the dark knowledge distillation: the teacher model is firstly trained
and then is used to generate targets for the training data. These targets are actually posterior probabilities and so are
`soft' compared to the original one-hot `hard' targets. The soft targets are used to train the child model. As we will see,
using soft targets leads to a smoother objective function, which makes the pre-training a much easier task than training with
the original hard targets.

Our experiments on an ASR task with the Aurora4 database demonstrated three interesting findings: (1) the knowledge transfer pre-training
can be used to train RNNs, which is challenging with conventional methods; (2) the pre-training can use a very weak teacher
model; (3) combining the knowledge transfer pre-training  and the conventional RBM  pre-training delivers additional gains.

The reset of the paper is organized as follows. Section~\ref{sec:rel} briefly discusses some related works, and~\ref{sec:method} presents the knowledge transfer pre-training. Section~\ref{sec:exp} presents the experiments, and the paper is concluded by Section~\ref{sec:con}.

\section{Related to prior work}
\label{sec:rel}

This study is directly motivated by the work of dark knowledge distillation from Hinton~\cite{hinton2014distilling}. The important distinction
is that we use simple models to teach complex models. The teacher model in our work in fact knows not so much, but it is sufficient to provide a rough guide that is important to train complex models, such as highly deep DNNs or multi-layer RNNs. More precisely, the existing methods use the teacher model as a knowledge source, while our method uses the teacher model for pre-training.

Our work is also related to the FitNets approach proposed by Romero et al.~\cite{romero2014fitnets}, where a teacher model is used to
supervise the learning of another network which can be in a different structure, e.g., deeper and fatter. Particularly, they learned
hidden layers instead of output layers, which is a big advantage in transferring hierarchical knowledge into child models.
Our approach focuses on learning the output layer, which does not
consider the internal structure of the teacher model, and so is truly `blind learning'. This offers more flexibility to pre-train complex and heterogeneous models, though looses the advantage of learning hierarchical patterns.

Another related work is the HMM-based pre-training approach recently proposed by Pasa and colleagues~\cite{pasahmm}.
In that work, the authors train an HMM model, and then use the trained model to generate training
data. The generated data are then used  to pre-train RNN models. This approach shares the same idea
of knowledge transfer pre-training as our work. The main difference is that the knowledge transfer in Pasa's approach is based on
some randomly sampled data, which essentially simulates the joint distribution of the data and their target labels;
whereas our approach is based on targets predicted by the teacher model, which simulates
the conditional distribution of the targets given the data.


\section{Pre-training with dark knowledge transfer}
\label{sec:method}

\subsection{Dark knowledge distiller}

The idea that a well-trained DNN model can be used as a teacher to help training other models was proposed by Ba and Hinton~\cite{ba2014deep,hinton2014distilling,li2014learning}. The basic assumption is that the teacher model
learns rich knowledge from the training data and this knowledge can be used to guide the training of child models
which are simple and hence unable to learn many details without the teacher's guide. To distill the knowledge from
the teacher model,
the logit matching approach proposed by Ba~\cite{ba2014deep} teaches the child model by encouraging
its logits (activations before softmax) close to those generated by the teacher model in terms of square error, and the dark
knowledge distiller model proposed by Hinton~\cite{hinton2014distilling} encourages the output
of the child model close to those of the teacher model in terms of cross entropy. This knowledge transfer
idea has been applied to learn simple models from complex models so that the simple model can
approach the performance of the complex model~\cite{li2014learning,chan2015transferring}.

We focus on the dark knowledge distiller model rather than logit matching as it showed better performance in our experiments.
This model uses a well-trained DNN as the teacher model to predict the targets of the
training samples, and these targets are used to train
the child model. The predicted targets are actually posterior probabilities
of the targets associated with the DNN output, and they are soft because the class identities
with these targets are not as deterministic as with the original one-hot hard targets.
To make the targets softer, a temperature $T$ was introduced in~\cite{hinton2014distilling} to
scale the logits. This is formulated by $p_i = \frac{e^{z_i/T}}{\sum_j e^{z_j/T} }$,
where $i,j$ indexes the target classes.
As argued by Hinton~\cite{hinton2014distilling}, a larger $T$ allows
more information of non-targets to be distilled.

\subsection{Knowledge transfer pre-training}

In the original proposal~\cite{hinton2014distilling}, knowledge transfer was used to
train simple models with a complex model, and the goal is to achieve a
light-weighted model that can approach to the performance of the complex
model. We argue in this paper that knowledge transfer is a general
method and can be used to pre-train complex models with a simple model.

The basic assumption is that soft targets lead to a smoother objective function,
and so training with them is easier than training with the original hard targets.
Intuitively, soft targets offer probabilistic class labels which are not
as deterministic as hard targets. This matches the real situation where
uncertainty always exists in classification tasks. For example, in speech
recognition, it is often difficult to identify the phone class of a frame
due to the effect of co-articulation. Moreover, the uncertainty associated
with soft targets blurs the decision boundary of correct and incorrect targets.
The smoothness associated with soft targets has been stated in~\cite{hinton2014distilling},
where it was argued that soft targets result in less variant
gradients between training samples. This is equal to say that
the objective function is smooth. A smooth objective function is certainly
much easier to optimize, and in the case where the targets are
extremely soft (i.e., $T$ goes to infinity), the objective function
becomes flat and the optimization is trivial.

The ease of training with soft targets can be used to simplify training
complex models. Generally speaking, complex models (e.g., very deep or
with recurrent connections) involve a large amount of parameters
or complex dependencies among variables, which leads to twisted
objective functions that are hard to optimize~\cite{bengio1994learning,sutskever2013importance}.
To solve the problem, conventional layer-wise pre-training breaks
a complex model to simpler models that can be easily trained
individually, and then stack them back to form the complex model. The smoothness
on objective functions offered by knowledge transfer learning in the form of soft targets
provides a different way to simplify complex model training: instead of
breaking the complex model into simple models, we replace the twisted
objective function with a smoother one by using soft targets when training
the model.
By this approach, the difficulty in complex model training is greatly reduced, and the optimization
can be conducted on the entire model instead of a single layer as in layer-wise
pre-training. As long as the smoothed objective function possesses
a similar trend as the original objective function in gradients, training with the smoothed function would
result in a good initialization for the model parameters.

Note that learning soft targets is not the ultimate goal of the model training,  so
a fine-tuning step is required to refine the model with the original hard targets. In this sense,
the knowledge transfer learning is a pre-training step, which initializes the model
parameters in such a way that the fine-tuning has a good starting point to
reach a better local minimum, compared to training with hard targets from the beginning.


The knowledge transfer pre-training is related to the curriculum training approach discussed
in~\cite{romero2014fitnets}, where training samples that are easy to learn are
firstly selected in model training, while more difficult samples are selected
later when the model is strong enough. In knowledge transfer pre-training,
the soft targets can be regarded as easy samples and so are firstly used (in pre-training),
and hard targets are difficult samples and are used later (in fine-tuning).

We highlight that for knowledge transfer pre-training, the teacher model
is not necessarily very strong. The goal of the pre-training is to provide a
good initialization for fine-tuning, instead of knowledge transfer from one model to another,
so a model with reasonable quality is sufficient to be a teacher, although more
intelligent teachers are generally welcome.

\subsection{Comparison with layer-wise pre-training}

A confirmed advantage of layer-wise pre-training is that it can discover hierarchical patterns of
the input signal by unsupervised learning. This hierarchical patterns discovering is desirable for
several reasons: it is consistent with the information processing strategy
in human brains, and it can find invariant high-level features that are robust against noise and corruption.
A potential problem of layer-wise pre-training, however, is that the patterns are learned in an unsupervised
fashion, which means that they are purely derived by statistics without considering the task in hand.
For example in speech recognition, less frequently occurred patterns such as rare consonant phones
are difficult to discover, however they are important for the recognition tasks.

The knowledge transfer pre-training, on the other hand, is purely supervised and so it is
high greedy towards the target task. Additionally, this approach pre-trains the
entire model and so tends to be fast. Finally, it can pre-train models without clear multi-layer structures.
The disadvantage is, it is just a functional mimic to the teacher model
without considering any internal structure of the teacher model. Therefore,
it can neither discover any hierarchical patterns, nor learn them from the teacher model.

An interesting idea is to combine different types of pre-training methods.
For example, we can use layer-wise pre-training to discover hierarchical patterns, and then use knowledge transfer
pre-training to promote the patterns that are most important to the task. A simple approach
investigated in this paper is to employ the RMB pre-training and the knowledge transfer pre-training
sequentially, so that the advantages of both methods are leveraged.

\section{Experiments}
\label{sec:exp}

The proposed knowledge transfer pre-training is applied to train acoustic models for ASR systems. In the first experiment,
the knowledge transfer pre-training is used to train RNNs with a DNN as the teacher model. In the second experiment,
the knowledge transfer pre-training is compared with RBM pre-training and layer-by-layer supervised pre-training, and the combination
of knowledge transfer pre-training and RBM pre-training is also investigated.

\subsection{Data and baseline}

The experiments are conducted on the Aurora4 database in noisy conditions, and the data profile is largely standard: $7137$ utterances for model training,  $4620$ utterances for development and $4620$ utterances for testing. The Kaldi toolkit\footnote{http://kaldi.sourceforge.net/} is used to conduct the model training and performance evaluation, and the process largely follows the Aurora4 s5 recipe for GPU-based DNN training. Specifically, the training starts from constructing a system based on Gaussian mixture models (GMMs) with the standard $13$-dimensional MFCC features plus the first- and second-order derivatives. A DNN system is then trained with the alignment provided by the GMM system. The feature used for the DNN system is the $40$-dimensional Fbanks. A symmetric $11$-frame window is applied to concatenate neighboring frames, and an LDA transform is used to reduce the feature dimension to $200$, which forms the DNN input. The DNN architecture involves $4$ hidden layers and each layer consists of $2048$ units. The output layer is composed of $2008$ units, equal to the total number of Gaussian mixtures in the GMM system. The cross entropy is used as the training criterion, and the stochastic gradient descendent (SGD) algorithm is employed to perform the training.

\subsection{Knowledge transfer pre-training for RNN}

To train the RNN acoustic models, the DNN model of the baseline system is used as the teacher model.
The RNN is based on the LSTM structure, where the input features are the $40$-dimensional Fbanks, and the output units correspond
to the Gaussian mixtures as in the DNN model. The momentum is empirically set to $0.9$, and the starting learning rate is set to $0.0001$ by default.

The experimental results are reported in Table~\ref{tab:res}. The performance is evaluated in terms of two criteria: the frame accuracy (FA) and the word error rate (WER). While FA is more related to the training criterion (cross entropy), WER is more important for speech recognition. In Table~\ref{tab:res}, the FAs are reported on both the training set (TR FA) and the cross validation set (CV FA), and the WER is reported on the test set.

In Table~\ref{tab:res}, `RNN [raw]' is the RNN baseline trained with hard targets directly. `RNN [prt.]' denotes systems after knowledge transfer pre-training,
and `RNN [prt.+ft.]' denotes systems with both knowledge transfer pre-training and fine-tuning. Two settings of the temperature ($T$) are evaluated ($T$=1 and $T$=2),
and the performance with one and two LSTM layers are reported respectively.

\begin{table}[!htb]
\caption{Results with RNN Models}
\label{tab:res}
\centering
\begin{tabular}{l|c|c|c|c|c}
\hline
               & \# LSTM  &  T   & TR FA\% &CV FA\%   & WER\% \\
\hline
DNN [4 hidden layers] & 0   &  -   & 63.1 & 45.2   & 11.40 \\

\hline
RNN [raw]            & 1  & -    & 67.3 & 51.9   & 13.57 \\
RNN [prt.]           & 1  & 1    & 59.4 & 49.9   & 11.46 \\
RNN [prt.+ft.]       & 1  & 1    & 65.5 & 54.2   & 10.71 \\
RNN [prt.]           & 1  & 2    & 58.2 & 49.5   & 11.32 \\
RNN [prt.+ft.]       & 1  & 2    & 64.6 & 54.1   & 10.57 \\
\hline
RNN [raw]            & 2  & -    & 68.8 & 53.2   & 12.34 \\
RNN [prt.]           & 2  & 1    & 60.4 & 50.6   & 11.11 \\
RNN [prt.+ft.]       & 2  & 1    & 66.6 & 55.4   & 10.13 \\
RNN [prt.]           & 2  & 2    & 58.6 & 49.7   & 11.26 \\
RNN [prt.+ft.]       & 2  & 2    & 65.8 & 55.2   & {\bf 10.10} \\
\hline
\end{tabular}
\end{table}

From the results, it can be observed that the RNN baseline (RNN [raw]) can not beat the DNN baseline in terms of WER, although much effort has been devoted to calibrate the training process, including various trials on different learning rates and momentum values. This is consistent with the results published with the Kaldi recipe. Note that this does not mean RNNs are inferior to DNNs. From the FA results, it is clear that the RNN models are better in terms of frame accuracy. Unfortunately, this advantage is not propagated to the WER results on the test set. Additionally, the results shown here can not be interpreted as that RNNs are not suitable for ASR (in terms of WER). In fact several researchers have reported better WERs with RNNs than with DNNs, e.g.,~\cite{graves2014towards,sak2014long,weng2014recurrent}. Our results just say that with the Aurora4 database, the RNNs with the \emph{basic} training method do not generalize well in terms of WER.

This problem can be largely solved by the knowledge transfer pre-training. It can be seen from Table~\ref{tab:res} that with the pre-training only, the RNN systems obtain equal or even better performance in comparison with the DNN baseline, which means that the knowledge learned by DNN helps the RNN models move out of bad local minima that are caused by the complex objective function. Paying attention to the FA results, it can be seen that the pre-training does not improve FAs on the training set, but better FAs on the CV set and better WERs on the test set are obtained. This indicates that the pre-training leads to models that are more generalizable with respect to both datasets and evaluation metrics. After the fine-tuning with hard targets, the performances of RNN systems are significantly improved. Additionally, it can be found that a larger $T$ leads to worse FAs on both the training and CV datasets, but better WERs on the test dataset. This indicates that knowledge transfer pre-training contributes by delivering a more generalizable model instead of a more optimized model.

When comparing the RNNs that involve one and two LSTM layers, it can be found that the two layers of LSTMs deliver better performance.
Note that two layers of
LSTMs are rather complex in structure, and so pre-training it with layer-wise unsupervised models (e.g., RMBs) is rather difficult.  With knowledge transfer, the pre-training is rather simple.

\begin{table}[!htb]
\caption{RNN Results with a Weak DNN for Pre-Training}
\label{tab:res-single}
\centering
\begin{tabular}{l|c|c|c|c}
\hline
                    & T              & TR FA\% &CV FA\%   & WER\% \\
\hline
DNN [1 hidden layer]      & -              & 61.2 & 42.5   & 13.10 \\
RNN [raw]           & -              & 68.8 & 53.2   & 12.34  \\
\hline
RNN [prt.]          & 1              & 57.2 & 48.6   & 12.69 \\
RNN [prt.+ft.]      & 1              & 65.3 & 55.0   & 10.72 \\
RNN [prt.]          & 2              & 54.4  & 46.8  & 13.20   \\
RNN [prt.+ft.]      & 2              & 64.7  & 54.7  & {\bf 10.60}	 \\
\hline
\end{tabular}
\end{table}

Another interesting investigation is to use a very weak teacher model to conduct the pre-training. A DNN with only 1 hidden layer of 2048 units (not deep actually)
is trained and used as the teacher model. This model is much weaker than the DNN baseline which involves 4 hidden layers. The results presented in Table~\ref{tab:res-single} show that even with the weak model, the
pre-training works fairly well, although not as well as with the original strong teacher model. This results confirm our conjecture that
the teacher model is not necessarily very strong.  The principle role of the teacher is not to teach all the details
to the student, but a correct direction with which the student can learn by itself.

\subsection{Comparison of pre-training methods}

Compared to RNN, the DNN model is much simpler. After the extensive research in recent years, training DNNs is not a problem any more. For example in
speech recognition, training a DNN model with more than $5$ layers is rather simple even without any pre-training techniques~\cite{seide2011feature}.
In this experiment, we apply various pre-training methods to train DNN models. The goal is not to demonstrate the necessity of pre-training
in DNN model training, but to compare different pre-training approaches.

For a better comparison, we use a new DNN baseline which involves $4$ hidden layers and each layer involves $1024$ units (it was $2048$ in the RNN experiment);
further more, no LDA was employed for the input feature. This setting makes the training a little difficult as less hidden units need to find
the most discriminative input from larger feature vectors. With the original DNN baseline, the pre-training methods didn't show much help,
particularly with the layer-wise methods.

\begin{table}[!htb]
\caption{DNN Results with Various Pre-Training Methods}
\label{tab:res-com}
\centering
\begin{tabular}{l|c|c|c}
\hline
                                 & TR FA\% &CV FA\%   & WER\% \\
\hline
DNN-4H [4 hidden layers]         & 57.3 & 44.1   & 12.22  \\
DNN-1H [1 hidden layer]          & 54.8 & 41.7   & 13.90  \\
\hline
RBM                              & 58.1 & 45.8   & 11.42  \\
Layer-by-layer Discriminative    & 61.1 & 43.8   & 12.16  \\
Knowledge Transfer (DNN-4H)      & 60.0 & 45.6   & 11.43  \\
Knowledge Transfer (DNN-1H)      & 59.6 & 45.1   & 11.65  \\
\hline
RBM + Knowledge Transfer (DNN-4H)  & 59.5 & 46.2   & {\bf 11.13}  \\
RBM + Knowledge Transfer (DNN-1H)  & 59.4 & 46.1   & 11.25  \\
\hline
\end{tabular}
\end{table}

We compare three pre-training methods: the RBM-based pre-training~\cite{hinton2006reducing}, the layer-by-layer discriminative
pre-training~\cite{yu2010roles}, and the proposed knowledge transfer pre-training. For the knowledge transfer pre-training, two
teacher models are tested: one is the DNN baseline that involves 4 hidden layers (DNN-4H),
and the other is a simpler model with only 1 hidden layer (DNN-1H). Note that for
the knowledge transfer pre-training, the classification layer (the affine transform before softmax)
needs to be re-initialized randomly after pre-training, otherwise it would be difficult for
the fine-tuning to achieve reasonable improvement.

The results are reported in Table~\ref{tab:res-com}. It can be observed that the RBM pre-training and the knowledge transfer pre-training
(DNN-4H as the teacher model) achieve significant performance improvement, and their performance are rather similar ($11.42$ vs. $11.43$).
With the weak one-layer DNN as the teacher model, the result is slightly worse than with the four-layer DNN, but it is still
rather good. The layer-by-layer discriminative pre-training does not show much contribution in this experiment. These results demonstrate
that the knowledge transfer pre-training works at least as well as the state-of-the-art layer-wise pre-training methods.

Finally, the RBM approach and the knowledge transfer approach are combined, where the RMB approach conducts layer-by-layer unsupervised
pre-training and after that the knowledge transfer approach conducts supervised pre-training on the entire network. Fine tuning is finally conducted
to achieve the best model. The results are shown in Table~\ref{tab:res-com} as well. It can be seen that this combination leads to the
best performance ($11.13$) that we can obtain on this task. This demonstrates that the two pre-training methods are complementary, and
the combination can leverage their respective advantage.

\section{Conclusion}
\label{sec:con}

We proposed a novel pre-training approach based on knowledge transfer learning. Compared to conventional layer-wise
pre-training methods that initialize a complex network by stacking simple models layer by layer, the knowledge transfer pre-training conducts the
initialization by a smooth objective function. As a supervised pre-training it is more task-oriented, and as an entire-network
pre-training it is faster. The experimental results on the ASR task demonstrated that the new pre-training approach can
effectively help training complex models, even with a weak teacher model. For example, a DNN model has been successfully
used to pre-train an RNN model. Compared to RMB pre-training
and layer-by-layer discriminative pre-training, the new approach leads to comparable or even better performance. Additionally,
the RBM pre-training and the knowledge transfer pre-training can be combined, which has lead to additional performance gains
in our experiments. The future work involves studying knowledge transfer between heterogeneous models, e.g., from
probabilistic models to neural models.




\bibliographystyle{IEEEtran}
\small{
\bibliography{dark}
}

\end{document}